\documentclass[english,reprint, citeautoscript, aip, jcp]{revtex4-1}
\usepackage[T1]{fontenc}
\usepackage[utf8]{inputenc}
\usepackage{xcolor}
\usepackage{pdfcolmk}
\usepackage{array}
\usepackage{multirow}
\usepackage{amsmath}
\usepackage{amssymb}
\usepackage{graphicx}
\PassOptionsToPackage{normalem}{ulem}
\usepackage{ulem}

\makeatletter

\providecommand{\tabularnewline}{\\}
\providecolor{lyxadded}{rgb}{0,0,1}
\providecolor{lyxdeleted}{rgb}{1,0,0}
\usepackage{algorithm,algpseudocode}
\usepackage{hyperref}

\makeatother

\usepackage{babel}
\begin{document}

\title{Hierarchical modeling of molecular energies using a deep neural network}

\author{Nicholas Lubbers}
\email{nlubbers@lanl.gov}

\affiliation{Theoretical Division and CNLS, Los Alamos National Laboratory, Los
Alamos, New Mexico 87545, USA}

\author{Justin S. Smith}

\affiliation{Theoretical Division and CNLS, Los Alamos National Laboratory, Los
Alamos, New Mexico 87545, USA}

\affiliation{Department of Chemistry, University of Florida, Gainesville, Florida
32611, USA}

\author{Kipton Barros}

\affiliation{Theoretical Division and CNLS, Los Alamos National Laboratory, Los
Alamos, New Mexico 87545, USA}
\begin{abstract}
We introduce the Hierarchically Interacting Particle Neural Network
(HIP-NN) to model molecular properties from datasets of quantum calculations.
Inspired by a many-body expansion, HIP-NN decomposes properties, such
as energy, as a sum over hierarchical terms. These terms are generated
from a neural network—a composition of many nonlinear transformations—acting
on a representation of the molecule. HIP-NN achieves state-of-the-art
performance on a dataset of 131k ground state organic molecules, and
predicts energies with 0.26~kcal/mol mean absolute error. With minimal
tuning, our model is also competitive on a dataset of molecular dynamics
trajectories. In addition to enabling accurate energy predictions,
the hierarchical structure of HIP-NN helps to identify regions of
model uncertainty.
\end{abstract}
\maketitle

\section{Introduction}

Models of chemical properties have wide-ranging applications in fields
such as materials science, chemistry, molecular biology, and drug
design. Commonly, one treats the nuclei positions as fixed (the Born-Oppenheimer
approximation), and molecular properties follow from the quantum-mechanical
state of electrons. The many-body Schrödinger equation is extremely
difficult to solve fully, and in practice computational quantum chemistry
involves some level of approximation. Common choices are, e.g., Coupled
Cluster (CC)~\cite{Cizek66,Bartlett07} and Density Functional Theory
(DFT)~\cite{Kohn65,Engel11}. Such \emph{ab initio }methods typically
exhibit cubic or worse scaling in the number of electrons. Faster
calculations are crucial in contexts such as molecular dynamics (MD)
simulation or high-throughput molecular screening.

To improve efficiency, one may sacrifice accuracy. For example, the
effective interactions between nuclei may be modeled with local classical
potentials of fixed form. Such potentials may be parameterized to
match given experimental data or quantum\emph{ }calculations. Classical
potentials are extremely fast, and enable MD simulations of systems
with $10^{6}$–$10^{9}$ atoms. However, the parameterization process
is empirical and the resulting potentials may not transfer to new
systems or new dynamical processes. For example, it is notoriously
difficult to model the energetic barriers of bond breaking in a transferrable
way~\cite{Duin01,Senftle16}. Force fields are also known to lack
transferability to chemical environments that differ from those used
in the fitting process~\cite{Rauscher15}. One may also compromise
between\emph{ ab initio }and empirical methodologies; e.g., Density
Functional Tight Binding~\cite{Elstner98,Elstner14} enables MD simulations
of $10^{3}$–$10^{5}$ atoms~\cite{Mniszewski15}, but brings its
own challenges in parameterization and transferability.

Recently there has been tremendous interest in using machine learning
(ML) to automatically construct potentials based upon large datasets
of quantum calculations~\cite{Rupp12,Montavon12,Bartok13,Lilienfeld15,Hansen15,Shapeev15,Ferre16,De16,Huo17,Faber17,Artrith17,Gubaev17}.
This approach aims for the best of both worlds: the accuracy of\emph{
}full quantum calculations and efficiency comparable to empirical
classical potentials. An especially promising direction builds upon
recent advances in computer vision~\cite{Krizhevsky12,Simonyan14,LeCun15,He16}.
Convolutional neural nets are designed for translation-invariant processing
of an image plane via convolutional filters. Similar architectural
principles allow us to design neural nets that process molecules while
respecting translation, rotation, and permutation invariances~\cite{Behler07,Duvenaud15,Kearnes16}.
Modern neural net architectures automatically learn representations
of local atomic environments without requiring any feature engineering,
and achieve state of the art performance in predictions of molecular
properties~\cite{Han17,Gilmer17,Schuett17,Schuett17a}. An advantage
of neural nets (compared to, e.g., kernel ridge regression and Gaussian
process regression~\cite{Rupp15,Bartok15}) is that the training
time scales linearly in the number of data points, making it practical
to learn from databases of millions of quantum calculations~\cite{Smith17}.

In this paper, we introduce the Hierarchically Interacting Particle
Neural Network (HIP-NN), which takes inspiration from the many-body
expansion (MBE). Following common practice~\cite{Bartok15,Rupp15,Behler15},
we assume that the \emph{ab initio} total energy $E$ of a molecule
may be modeled as a sum over local contributions at each atom $i$,
\begin{equation}
E\approx\hat{E}=\sum_{i=1}^{N_{\mathrm{atom}}}\hat{E}_{i}.\label{eq:energy_1}
\end{equation}
HIP-NN further decomposes the local energy model $\hat{E}_{i}$ in
contributions over orders $n$,
\begin{equation}
\hat{E}_{i}=\sum_{n=0}^{N_{\mathrm{interaction}}}\hat{E}_{i}^{n}.\label{eq:energy_2}
\end{equation}
The MBE, commonly employed in classical potentials~\cite{Stillinger85,Elrod94},
would use $\hat{E}^{n}$ to represent $(n+1)$-body contributions
to the energy, i.e., interactions between atom $i$ and up to $n$
of its neighbors. Integration of the MBE into ML models of molecular
energies has been suggested in Refs.~\onlinecite{Bartok15,Yao17}.
This prior work employed separate\emph{ }ML models for each expansion
order $n$. A key aspect of HIP-NN is that a \emph{single} network
produces $\hat{E}_{i}^{n}$ at all orders, allowing these terms to
be simultaneously learned in coherent way. Furthermore, the HIP-NN
ansatz is more general than the MBE, in that the terms $E_{i}^{n}$
may incorporate many-body interactions at higher order than $n$.
The decomposition is non-unique, but should be designed such that
$\hat{E}_{i}^{n}$ rapidly vanishes with increasing order $n$. To
pursue this, our training procedure utilizes a hierarchical regularization
term to encourage the outputs $\hat{E}_{i}^{n}$ to decay with $n$.
After training, if decay with $n$ is \emph{not} observed for a given
input molecule, then the HIP-NN energy prediction is less likely to
be accurate. That is, HIP-NN can estimate the reliability of its own
energy predictions. 

We detail the HIP-NN architecture and training procedure in the next
section. Section~\ref{sec:results} demonstrates that HIP-NN effectively
learns molecular energies for various benchmark datasets. On the QM9
dataset of organic molecules~\cite{Ramakrishnan14}, HIP-NN predicts
energies with a ground-breaking mean absolute error of 0.26~kcal/mol.
HIP-NN also performs well on datasets of MD trajectories with minimal
tuning. Variants of HIP-NN achieve good performance with parameter
counts ranging from $\sim10^{3}$ to $10^{5}$. In addition to enabling
robust predictions, the hierarchical structure of HIP-NN provides
a built-in measure of model uncertainty. In Sec.~\ref{sec:discussion}
we further discuss and interpret our numerical results, and we conclude
in Sec.~\ref{sec:conclusion}.

\section{HIP-NN methodology\label{sec:method}}

\begin{figure}
\includegraphics[clip,width=0.8\columnwidth]{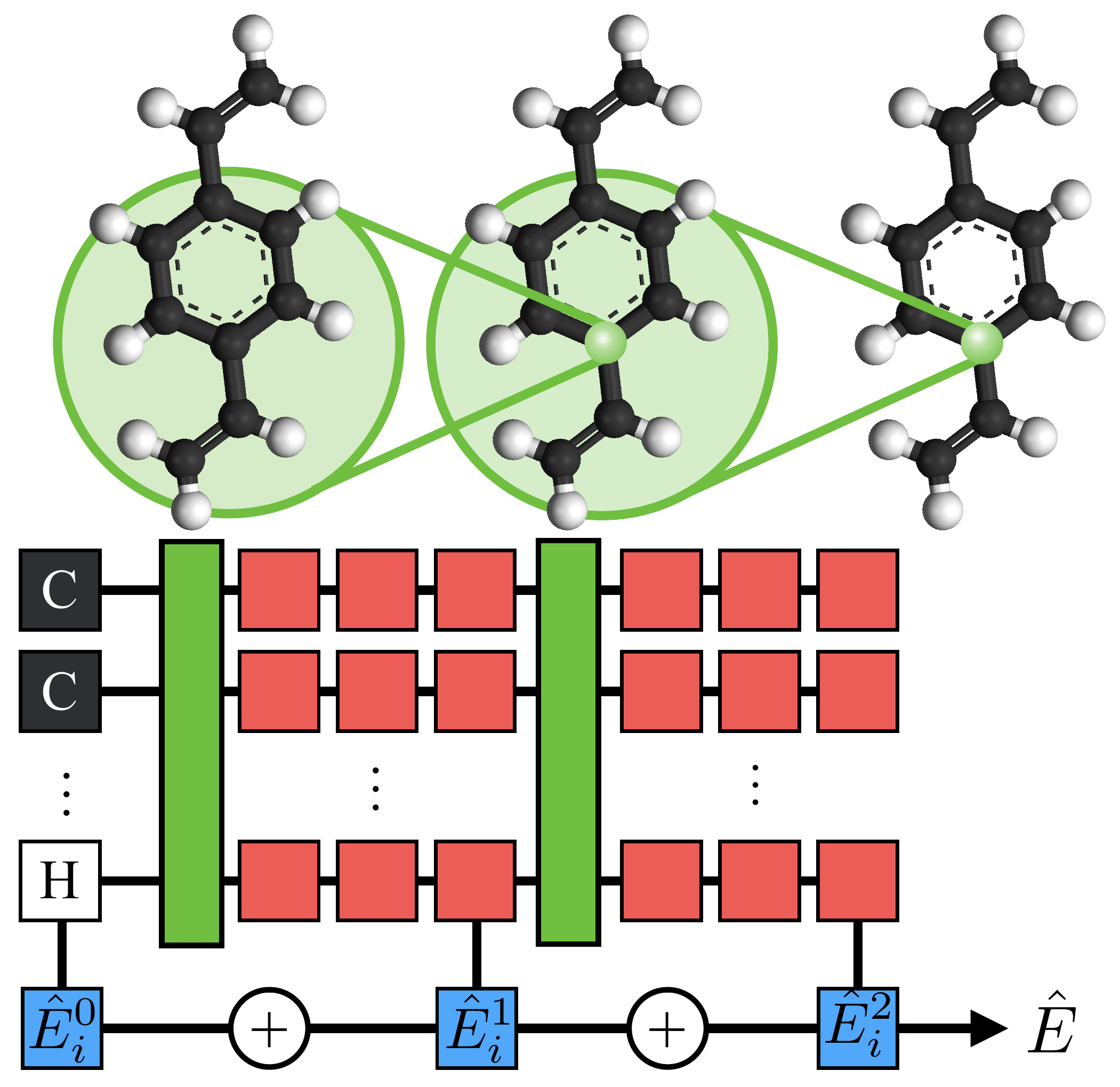}

\caption{\label{fig:schematic}HIP-NN processes a molecule from left to right,
building successive \emph{atomic features} to describe the local chemical
environments. This processing occurs through \emph{interaction }and
\emph{on-site }layers\emph{ }(green and red boxes, respectively).
Interaction layers collect information from a local neighborhood (green
circles). The total molecular energy $\hat{E}$ includes contributions
$\hat{E}_{i}^{n}$ at all sites $i$ and hierarchical levels $n$. }
\end{figure}
 Figure~\ref{fig:schematic} illustrates HIP-NN, our neural network
for predicting molecular properties and energies. The molecular configuration
is input on the left using a simple representation, discussed below,
that is symmetric with respect to translation, rotation, and permutation
of atoms. As the molecule is processed from left to right, HIP-NN
builds consecutive sets of \emph{atomic features }to characterize
the chemical environment of each atom. Blue boxes denote hierarchical
contributions to the total energy—the final output of HIP-NN. Green
boxes denote \emph{interaction }layers, which mix information between
pairs of atoms within some radius (illustrated for a single carbon
atom using green circles). Red boxes denote \emph{on-site }layers\emph{,
}which process the atomic features of a single atom. These components
are described mathematically in the subsections below.

\subsection{Molecular representation}

A molecular configuration $\mathcal{C}=\{(Z_{i},\mathbf{r}_{i})\}$
is defined by the atomic numbers $Z_{i}$ and coordinates $\mathbf{r}_{i}$
of atoms $i=1\dots N_{\mathrm{atom}}$. We seek a representation of
$\mathcal{C}$ suitable for input to HIP-NN. 

To achieve a representation of the molecular geometry that is invariant
under rigid transformations (i.e., translations, rotations, and reflections)
we work with pairwise distances $r_{ij}=|\mathbf{r}_{i}-\mathbf{r}_{j}|$
rather than coordinates $\mathbf{r}_{i}$. Furthermore, we keep only
distances satisfying $r_{ij}<R_{\mathrm{cut}}$. In our energy model,
we apply a smooth radial cutoff to ensure smoothness with respect
to atomic positions.

We represent the atomic numbers $Z_{i}$ using a one-hot encoding,

\begin{equation}
z_{i,a}^{0}=\delta_{Z_{i},\mathcal{Z}(a)},\label{eq:one_hot}
\end{equation}
where $\delta_{ij}$ is the Kronecker delta and $\mathcal{Z}$ enumerates
the atomic numbers under consideration. We benchmark on datasets of
organic molecules containing atomic species {[}H, C, N, O, F{]} for
which $\mathcal{Z}=[1,6,7,8,9]$. By construction, HIP-NN will sum
over atomic and feature indices ($i$ and $a$), and is thus invariant
to their permutation.

\subsection{Atomic features and energies}

HIP-NN generalizes $z_{i,a}^{0}$ to real-valued, dimensionless \emph{atomic
features} $z_{i,a}^{\ell}$ (i.e., neural network activations) over
\emph{layers} indexed by $\ell=0\dots N_{\mathrm{layer}}$~\cite{Behler07}.
Suppressing the feature index $a=1\dots N_{\mathrm{feature}}^{\ell}$,
we call $z_{i}^{\ell}$ the feature vector. The input feature vector
$z_{i}^{0}$ represents the species of atom $i$. At subsequent layers,
HIP-NN generates successively more abstract, ``dressed'' representations
$z_{i}^{\ell+1}$ of the chemical environment of atom $i$ based upon
information $(z_{j}^{\ell},r_{ij})$ from neighboring atoms $j$.

The key challenge in HIP-NN is to learn good features $z_{i,a}^{\ell}$
that faithfully capture the chemical environment of atom $i$. Once
known, HIP-NN uses linear regression (blue boxes in Fig.~\ref{fig:schematic})
on the atomic features to model the local hierarchical energies,
\begin{align}
\hat{E}_{i}^{n} & =\sum_{a=1}^{N_{\mathrm{feature}}}w_{a}^{n}z_{i,a}^{\ell_{n}}+b^{n},\label{eq:ehat_n}
\end{align}
where $w_{a}^{n}$ and $b^{n}$ are learned parameters with dimensions
of energy. The total HIP-NN energy $\hat{E}$ is then given by Eqs.~(\ref{eq:energy_1})
and~(\ref{eq:energy_2}). Note that only certain network layers $\ell_{n}$
contribute to the energy.

\subsection{On-site layers}

The on-site layers (red squares in Fig.~\ref{fig:schematic}) operate
on the features $z_{i,a}^{\ell}$ of a single atom,

\begin{equation}
\tilde{z}_{i,a}^{\ell+1}\underset{\mathrm{\textrm{on-site}}}{=}f\left(\sum_{b}W_{ab}^{\ell}z_{i,b}^{\ell}+B_{a}^{\ell}\right),\label{eq:local_layer}
\end{equation}
where $W_{a,b}^{\ell}$ and $B_{a}^{\ell}$ are learned parameters.
Various choices of \emph{activation function} $f(x)$ are possible.
Rectifiers (i.e., functions saturating for $x\rightarrow-\infty$
and increasing indefinitely when $x\rightarrow\infty$) are often
preferred because they help mitigate the so-called vanishing gradient
problem~\cite{Hochreiter01,Glorot11}. For HIP-NN, we select the
softplus activation function~\cite{Dugas01,Schuett17a},
\begin{equation}
f(x)=\log(1+e^{x}).\label{eq:softplus}
\end{equation}

To obtain the final atomic features at layer $\ell+1$, we apply a
residual network (ResNet) transformation~\cite{He16},

\begin{equation}
z_{i,a}^{\ell+1}=\sum_{b}\left(\tilde{W}_{ab}^{\ell}\tilde{z}_{b}^{\ell+1}+\tilde{M}_{ab}^{\ell}z_{i,b}^{\ell}\right)+\tilde{B}_{a}^{\ell},\label{eq:resnet}
\end{equation}
where $\tilde{W}_{ab}^{\ell}$, $\tilde{M}_{ab}^{\ell}$, and $\tilde{B}_{a}^{\ell}$
are again learned parameters. Following the suggestion of the ResNet
authors, if layers $\ell$ and $\ell+1$ have the same number of features,
we instead make $\tilde{M}_{ab}$ unlearnable, and fix $\tilde{M}_{ab}^{\ell}=\delta_{ab}$.
Empirically, the ResNet architecture further mitigates the vanishing
gradients problem, allowing training of deeper networks.

\subsection{Interaction layers}

\begin{figure}
\includegraphics[width=0.9\columnwidth]{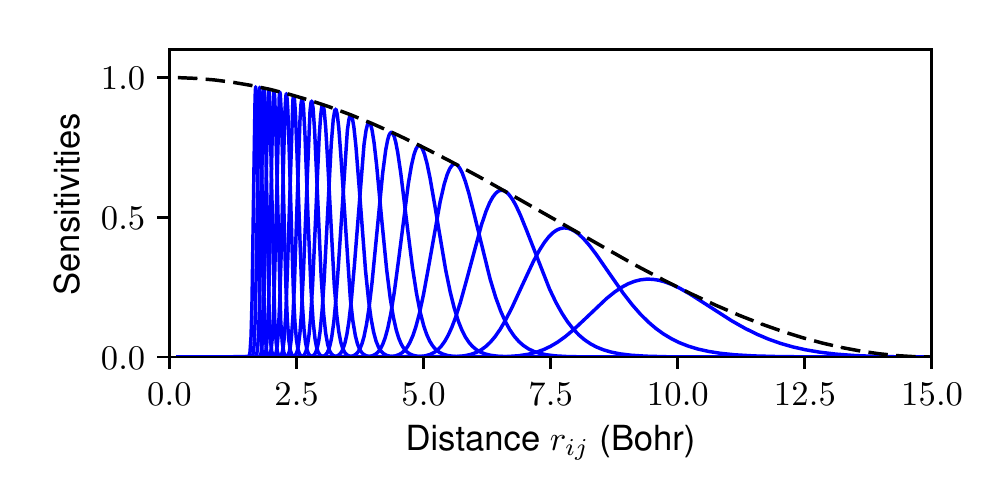}\caption{\label{fig:sensitivity}Within an interaction layer, the spatial sensitivity
functions $s_{\nu}(r_{ij})$ modulate communication between pairs
of atoms separated by distance $r_{ij}$. Blue curves: The initial
sensitivities $s_{\nu}(r_{ij})$ for $\nu=1\dots20$. Dashed black
curve: All sensitivities are scaled by the factor $\varphi_{\mathrm{cut}}(r_{ij})$,
which introduces a hard spatial cutoff of 15 Bohr. }
\end{figure}

The interaction layers (green boxes in Fig.~\ref{fig:schematic})
operate similarly to on-site layers, Eq.~(\ref{eq:local_layer}),
and additionally transmit information between atoms. The transformation
rule for interaction layers is

\begin{equation}
\tilde{z}_{i,a}^{\ell+1}\underset{\mathrm{inter.}}{=}f\left(\sum_{j,b}v{}_{ab}^{\ell}(r_{ij})z_{j,b}^{\ell}+\sum_{b}W_{ab}^{\ell}z_{i,b}^{\ell}+B_{a}^{\ell}\right),\label{eq:inter_layer}
\end{equation}
where $v_{ab}^{\ell}(r_{ij})$ collects information from neighboring
atoms $j$ that are sufficiently close to $i$, i.e., that satisfy
$r_{ij}<R_{\mathrm{cut}}$. We expand the $r_{ij}$ dependence in
the basis of \emph{sensitivity }functions,
\begin{equation}
v_{ab}^{\ell}(r_{ij})=\sum_{\nu}V_{\nu,ab}^{\ell}s_{\nu}^{\ell}(r_{ij}),\label{eq:inter_tensor}
\end{equation}
with learned parameters $V_{\nu,ab}^{\ell}$. We select the spatial
sensitivities to be Gaussian in inverse distance,
\begin{equation}
s_{\nu}^{\ell}(r)=\exp\left[-\frac{\left(r^{-1}-\mu_{\nu,\ell}^{-1}\right)^{2}}{2\sigma_{\nu,\ell}^{-2}}\right]\varphi_{\mathrm{cut}}(r).\label{eq:sensitivity}
\end{equation}
The distances $\mu_{\nu,\ell}$ and $\sigma_{\nu,\ell}$ are learned
parameters. We modulate the sensitivities with the cutoff function,
\begin{equation}
\varphi_{\mathrm{cut}}(r)=\begin{cases}
\left[\cos\left(\frac{\pi}{2}\frac{r}{R_{\mathrm{cut}}}\right)\right]^{2} & r\le R_{\mathrm{cut}}\\
0 & r>R_{\mathrm{cut}}
\end{cases}.
\end{equation}
Figure~\ref{fig:sensitivity} illustrates the sensitivity and cutoff
functions for the initial parameters described in Sec.~\ref{subsec:initial}.

Interaction and on-site layers use the same activation function, Eq.~(\ref{eq:softplus}),
and ResNet transformation, Eq.~(\ref{eq:resnet}).

\subsection{Training}

\subsubsection{Loss function}

The goal is to accurately predict molecular properties. We evaluate
both the Mean Absolute Error and Root-Mean-Square Error,
\begin{align}
\mathrm{MAE} & =\langle|(\hat{E}-E)|\rangle_{D},\\
\mathrm{RMSE} & =\sqrt{\langle(\hat{E}-E)^{2}\rangle_{D}}.
\end{align}
The brackets $\langle\cdot\rangle_{D}$ denote an average over molecules
within a dataset $D$, $\hat{E}$ is the molecular energy predicted
in Eq.~(\ref{eq:energy_1}), and $E$ is the true \emph{ab initio}
energy.

We optimize the HIP-NN model parameters to minimize both MAE and RMSE.
That is, we wish to minimize a loss function,
\begin{equation}
\mathcal{L}=\frac{1}{\sigma_{E}}(\mathrm{MAE}+\mathrm{RMSE})+\mathcal{L}_{L2}+\mathcal{L}_{R}.\label{eq:loss}
\end{equation}
In this context, we select $D=D_{\textrm{train}}$ to be the training
dataset. The natural energy scale for MAE and RMSE is the standard
deviation of molecular energies, 
\begin{equation}
\sigma_{E}=\sqrt{\langle(E-\langle E\rangle)^{2}\rangle_{D}}.\label{eq:sigma_e}
\end{equation}

Importantly, the loss function includes two regularization terms.
The first is a $L_{2}$ regularization on weight tensors appearing
in the equations for energy regression~(\ref{eq:ehat_n}), on-site
layers~(\ref{eq:local_layer}), interaction layers~(\ref{eq:inter_layer}),~(\ref{eq:inter_tensor}),
and ResNet transformation~(\ref{eq:resnet}):
\begin{equation}
\mathcal{L}_{L2}=\lambda_{L2}\left(\frac{||w||_{2}^{2}}{\sigma_{E}^{2}}+||W||_{2}^{2}+||V||_{2}^{2}+||\tilde{W}||_{2}^{2}+||\tilde{M}||_{2}^{2}\right).\label{eq:l2_regularization}
\end{equation}
We find that a sufficiently small hyperparameter $\lambda_{L2}$ is
effective at reducing outlier HIP-NN predictions while introducing
minimal bias to the model.

To encourage hierarchality of the energy terms, we include a second
regularization term,
\begin{equation}
\mathcal{L}_{R}=\lambda_{R}\langle R\rangle_{D},\label{eq:hierarchical_regularization}
\end{equation}
that penalizes the \emph{non-hierarchicality $R$} of energy contributions,
\begin{equation}
R=\sum_{n=1}^{N_{\mathrm{interaction}}}\sum_{i=1}^{N_{\mathrm{atom}}}\frac{(\hat{E}_{i}^{n})^{2}}{(\hat{E}_{i}^{n})^{2}+(\hat{E}{}_{i}^{n-1})^{2}}.\label{eq:hierarchicality}
\end{equation}
When HIP-NN is functioning properly, we commonly observe that $\hat{E}_{i}^{n}$
decays rapidly in $n$. A large value of $R$ thus indicates malfunction
of HIP-NN for the given molecular input.

\subsubsection{Stochastic optimization}

We use the Adaptive Moment Estimation (Adam) algorithm~\cite{Kingma14},
a variant of stochastic gradient descent (SGD), to train HIP-NN. Let
$U=\{w,b,W,B,V,\sigma,\mu,\tilde{W},\tilde{B},\tilde{M}\}$ denote
the full set of model parameters, Eqs.~(\ref{eq:ehat_n})–(\ref{eq:sensitivity}).
The goal, then, is to evolve $U$ to minimize the loss $\mathcal{L}\bigl|_{D}$,
Eq.~(\ref{eq:loss}), evaluated on the dataset $D=D_{\textrm{train}}$
of training molecules.

In SGD, one partitions the training data into random disjoint sets
of \emph{mini-batches}, $D=\tilde{D}_{1}\cup\tilde{D}_{2}\dots\cup\tilde{D}_{N}$.
For each mini-batch $\tilde{D}$, one evolves $U$ in the direction
of the negative gradient $-\nabla_{U}\mathcal{L}\bigl|_{\tilde{D}}$,
which is a stochastic approximation to $\nabla_{U}\mathcal{L}\bigl|_{D}$,
the gradient evaluated on the full dataset. Training time is measured
in \emph{epochs}. Each epoch corresponds to a pass through all mini-batches
$\tilde{D}_{i}$. After each epoch, the mini-batch partition is re-randomized.
Compared to plain SGD, Adam speeds convergence by selecting its updates
as a function of a decaying average of previous gradients. The Adam
parameters are its learning rate $\eta$ and exponential decay factors
$\beta_{1}$ and $\beta_{2}$.

To reduce overfitting, we use an \emph{early stopping} procedure to
terminate the learning process when the MAE on a \emph{validation}
dataset $D_{\mathrm{validate}}$ (separate from the training data
$D_{\textrm{train}}$) stops improving~\cite{Morgan90}. The Adam
learning rate $\eta$ is initialized to $\eta_{\textrm{init}}$ and
annealed as follows. We train the network while tracking \texttt{best\_score},
the best validation MAE yet observed, and corresponding model parameters
$U$. The learning rate is fixed to $\eta_{\textrm{init}}$ for the
first $t_{\textrm{init}}$ epochs. Afterwards, if \texttt{best\_score}
plateaus (does not drop for a period of $t_{\mathrm{patience}}$ epochs)
then the learning rate $\eta$ is multiplied by $\alpha_{\mathrm{decay}}$,
causing the gradient descent procedure to take finer steps. Training
is terminated if $\eta$ decreases twice without any improvement to
\texttt{best\_score}. Training is forcefully terminated if $t_{\textrm{max}}$
epochs elapse. The final parameter set $U$ is taken to be the one
which produced the lowest validation error.

\subsection{Implementation details\label{subsec:initial}}

Here we discuss hyperparameters, initialization of model parameters,
and our numerical implementation.

As illustrated in Fig.~(\ref{fig:schematic}) we use $n=0\dots N_{\textrm{interaction}}$
hierarchical contributions to the energy model. We choose $N_{\textrm{interaction}}=2$
interaction layers, a number comparable to previous studies~\cite{Duvenaud15,Schuett17,Gilmer17,Schuett17a}.
Each interaction layer is followed by $N_{\textrm{on-site}}$ on-site
layers. Thus the total number of nonlinear layers is $N_{\textrm{interaction}}\times(1+N_{\textrm{on-site}})$.
We fix the feature vector size to a constant $N_{\textrm{feature}}=|z_{i}^{\ell}|$
for all layers $\ell>0$. Recall that the input feature vector $z_{i}^{0}$
is a one-hot encoding of the atomic species. In our numerical studies,
we consider models with varying $N_{\textrm{on-site}}$ and $N_{\textrm{\textrm{feature}}}$
hyperparameters.

The initial network weights $w,$ $W$, $V$, $\tilde{W}$, and $\tilde{M}$
from Eqs.~(\ref{eq:ehat_n})–(\ref{eq:inter_tensor}) are drawn from
a uniform distribution according to the Glorot initialization scheme~\cite{Glorot10}.
We initialize the network biases $b$, $B$, and $\tilde{B}$ to zero.
Next, we set the zeroth-order energy model $\hat{E}^{n=0}$ to minimize
the least squares error on the training data. The corresponding linear
regression parameters $w_{a}^{0}$ and $b^{0}$ are held fixed for
the duration of training. For subsequent orders $n>0$ we rescale
the Glorot initialized weights $w_{a}^{n}$ by a factor $\sigma_{E}/10^{-2n}$
to impose the expected energy scale and hierarchical decay. During
training, we factorize $w_{a}^{n}=\sigma_{E}\tilde{w}_{a}^{n}$ and
treat $\tilde{w}_{a}^{n}$ as the learnable, dimensionless parameters.

We employ $N_{\textrm{sensitivity}}=20$ sensitivity functions $s_{\nu}^{\ell}(r)$
as given by Eq.~(\ref{eq:sensitivity}). Initially, the sensitivities
are independent of layer $\ell$. We select initial inverse distances
$\mu_{\nu,\ell}^{-1}$ with uniform separation between $R_{\textrm{low}}^{-1}$
and $R_{\textrm{high}}^{-1}$ for $\nu=1\dots N_{\textrm{sensitivity}}$.
The lower and upper distances are $R_{\textrm{low}}=1.7$ Bohr and
$R_{\textrm{high}}=10$ Bohr. The width parameters $\sigma_{\nu,\ell}$
are initialized to the constant value $2N_{\textrm{sensitivity}}R_{\textrm{low}}$,
which allows moderate overlap between adjacent sensitivity functions,
as shown in Fig.~\ref{fig:sensitivity}. The sensitivities are modulated
by a cutoff $R_{\textrm{cutoff}}=15$~Bohr.

The loss function regularization terms are weighted by $\lambda_{L2}=10^{-6}$
and $\lambda_{R}=10^{-2}$. The training data mini-batches each contain
$30$ molecules. An additional validation dataset of 1000 molecules
(separate from both training and test data) is used to determine the
early stopping time. The Adam decay hyperparameters are $\beta_{1}=0.9$,
$\beta_{2}=0.999$, and the initial learning rate is $\eta_{\textrm{init}}=10^{-3}$.
The learning rate decay factor is $\alpha_{\textrm{decay}}=0.5$.
The patience is $t_{\textrm{patience}}=50$ epochs, the initial training
time before annealing is $t_{\textrm{init}}=100$, and the maximum
training time is $t_{\textrm{max}}=2000$.

We implemented HIP-NN using the Theano framework~\cite{TDT16} and
Lasagne neural network library~\cite{Dieleman15} with custom layers.
Theano calculates gradients of the loss function using backpropagation
(also known as reverse-mode automatic differentiation~\cite{Griewank89}).
Theano also compiles the model for high performance execution on GPU
hardware. A single Nvidia Tesla P100 GPU requires about 1 minute to
complete one training epoch for the full QM9 dataset (discussed below)
with $N_{\textrm{on-site}}=3$ and $N_{\textrm{feature}}=80$. The
full training procedure typically completes in 1000 to 2000 epochs.

\section{Results}

\label{sec:results}

\subsection{QM9 dataset}

The QM9\emph{ }dataset~\cite{Ruddigkeit12,Ramakrishnan14} is comprised
of about $134$k organic molecules containing H and nine or fewer
C, N, O, and F atoms. Properties were calculated at the B3LYP/6-31G(2df,p)
level of quantum chemistry. About $3$k molecules within QM9 fail
a geometric consistency check~\cite{Ramakrishnan14} and are commonly
removed from the dataset~\cite{Gilmer17,Schuett17a,Faber17}. The
authors of QM9 had difficulty in the energy-minimization procedure
for 11 more molecules~\cite{Ramakrishnan14}, which we also remove.
Our pruned dataset thus contains about $131$k molecules. This dataset
is then randomly partitioned into training, validation, and testing
datasets, $D_{\textrm{all}}=D_{\textrm{train}}\cup D_{\textrm{validate}}\cup D_{\textrm{test}}$.
We benchmark on varying amounts of training data, $|D_{\textrm{train}}|=N_{\textrm{train}}$.
The validation dataset controls early stopping and has fixed size
$|D_{\textrm{validate}}|=1000$. All remaining molecules are included
in the testing dataset $D_{\textrm{test}}$. Every HIP-NN error statistic
$S$ reported below (e.g., MAE and RMSE over $D_{\textrm{test}}$)
is actually a sample average $\mu_{S}$ over $N_{\textrm{model}}=8$
models, each with a differently randomized split of the training/validation/testing
data. We calculate error bars as $\sigma_{S}/\sqrt{N_{\textrm{model}}}$,
where $\sigma_{S}$ is the sample standard deviation over the $N_{\textrm{model}}$
models.

\begin{table*}
\caption{\label{tab:QM9comparison}QM9 performance (MAE in kcal/mol) for various
models reported in the literature.}

\begin{tabular}{lllllll}
\hline 
$N_{\textrm{train}}+N_{\textrm{validate}}$ & HIP-NN & MTM~\cite{Gubaev17} & SchNet~\cite{Schuett17a} & MPNN~\cite{Gilmer17} & HDAD+KRR~\cite{Faber17} & DTNN~\cite{Schuett17} \tabularnewline
\hline 
\hline 
110426 & \textbf{$\mathbf{0.256}\pm0.003$} & ~~- & $0.31$ & $0.42$ & $0.58$ & ~~-\tabularnewline
100000 & $\mathbf{0.261}\pm0.002$ & ~~- & $0.34$ & ~~- & ~~- & $0.84$\tabularnewline
50000 & $\mathbf{0.354}\pm0.004$~~ & 0.41 & $0.49$ & ~~- & ~~- & $0.94$\tabularnewline
\hline 
\end{tabular}
\end{table*}

Table~\ref{tab:QM9comparison} benchmarks HIP-NN against recent state-of-the-art
models reported in the literature. The HIP-NN models contain $N_{\textrm{on-site}}=3$
on-site layers and $N_{\textrm{feature}}=80$ atomic features per
layer. Following previous work, we report the mean absolute error
(MAE) using training sets of three different sizes. HIP-NN achieves
an MAE of 0.26~kcal/mol when trained on the largest datasets and,
to our knowledge, outperforms all existing models.

\begin{table*}
\caption{\label{tab:QM9_hipnns}QM9 performance for HIP-NN models with varying
complexity.}
\begin{tabular}{lllll}
\hline 
$N_{\textrm{feature}}$ & Parameter count~~ & MAE (kcal/mol)~~ & RMSE (kcal/mol)~~ & Errors above 1 kcal/mol (\%)\tabularnewline
\hline 
\hline 
5 & 1.6k & $1.177\pm0.014$ & $1.851\pm0.019$ & $42.18\pm0.58$\tabularnewline
10 & 4.9k & $0.653\pm0.007$ & $1.077\pm0.015$ & $18.95\pm0.37$\tabularnewline
20 & 17k & $0.398\pm0.005$ & $0.706\pm0.025$ & $6.60\pm0.16$\tabularnewline
40 & 61k & $0.274\pm0.003$ & $0.539\pm0.014$ & $2.65\pm0.06$\tabularnewline
60 & 134k & $0.261\pm0.004$ & $0.552\pm0.024$ & $2.37\pm0.09$\tabularnewline
80 & 234k & $0.256\pm0.003$ & $0.527\pm0.020$ & $2.26\pm0.07$\tabularnewline
\hline 
60 (no hierarchy) & 134k & $0.278\pm0.008$ & $0.522\pm0.013$ & $2.75\pm0.21$\tabularnewline
80 (no hierarchy) & 234k & $0.293\pm0.008$ & $0.539\pm0.013$ & $3.10\pm0.21$\tabularnewline
\hline 
\end{tabular}
\end{table*}

Table~\ref{tab:QM9_hipnns} shows HIP-NN performance as a function
of model complexity. We fix $N_{\textrm{on-site}}=3$ on-site layers
and allow the number of atomic features $N_{\textrm{feature}}$ to
vary between 5 and 80. The HIP-NN parameter count grows roughly as
$N_{\textrm{feature}}^{2}$. For each complexity level we calculate
three error statistics: (1) MAE, (2) RMSE, and (3) the percentage
of molecules in the testing set whose predicted energy has an absolute
error that exceeds 1~kcal/mol (a common standard of chemical accuracy).
In the last two rows we report the performance of HIP-NN trained without
hierarchical energy contributions {[}i.e., fixing $w^{n}=b^{n}=0$
for $n=\{0,1\}$ in Eq.~(\ref{eq:ehat_n}), so that only $\hat{E}^{n=2}$
contributes to $\hat{E}${]}, and without hierarchical regularization,
Eq.~(\ref{eq:hierarchical_regularization}). With these limitations,
the MAE performance degrades by $\approx9\%$, the fraction of errors
above 1 kcal/mol increases by $\approx22\%$, but the RMSE values
are comparable.

Note that with only 5 atomic features (corresponding to 1.6k parameters)
the MAE of 1.2~kcal/mol already approaches chemical accuracy. This
performance is remarkable, given that the parameter count is roughly
two orders of magnitude smaller than the QM9 dataset size. For reference,
the standard deviation of energies in QM9 is $\sigma_{E}\approx238$~kcal/mol.
We observe that the HIP-NN error tends to decrease with increasing
$N_{\textrm{feature}}$, but the non-hierarchical HIP-NN model with
$N_{\textrm{feature}}=80$ performs worse than that with $N_{\textrm{feature}}=60$,
possibly due to overfitting.

Even though our best MAE of 0.26~kcal/mol is well under 1~kcal/mol,
approximately 2.3\% of the predicted molecular energies have an error
that exceeds 1~kcal/mol; there is still room for improved ML models
with fewer outliers in the energy predictions.

\begin{figure}
\includegraphics[width=0.9\columnwidth]{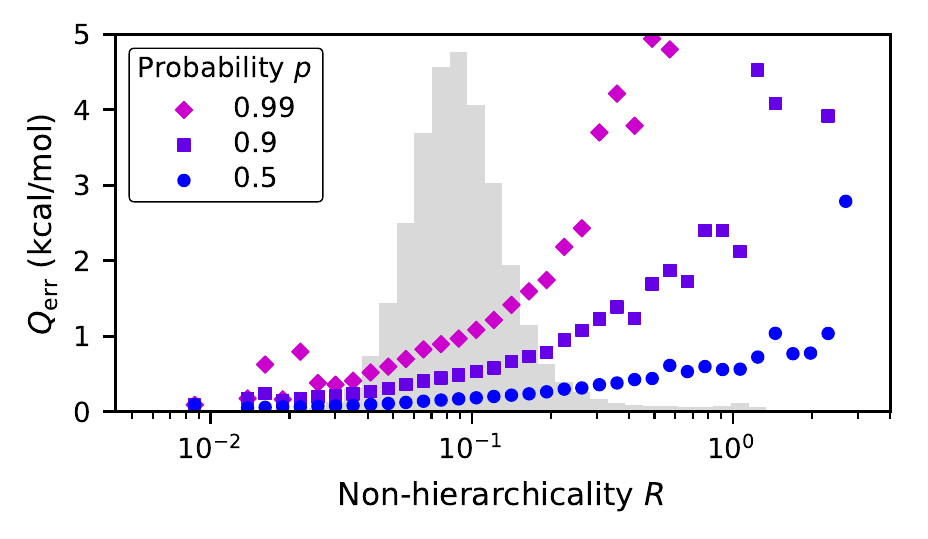}\caption{\label{fig:error_vs_reg}Larger non-hierarchicality $R$, Eq.~(\ref{eq:hierarchicality}),
indicates a breakdown of the energy hierarchy assumption, Eq.~(\ref{eq:energy_2}),
and correlates with larger error in the HIP-NN predictions, as observed
in quantile functions $Q_{\textrm{err}}(p,R)$ from Eq.~(\ref{eq:q_err}).
The gray background shows the rescaled probability distribution of
$\log\,R$. The scatter of $Q_{\textrm{err}}$ at very small and large
values of $R$ is likely due to a lack of data.}
\end{figure}

Figure~\ref{fig:error_vs_reg} shows that the non-hierarchicality
$R$ is an indicator of inaccurate HIP-NN energy predictions. This
is reasonable because large $R$ indicates breakdown of the energy
hierarchy assumption, i.e., non-decaying contributions $\hat{E}_{i}^{n}$
in Eq.~(\ref{eq:energy_2}). We quantify this corrrespondence by
considering the distribution of absolute error $|\hat{E}-E|$ over
the testing dataset $D_{\textrm{test}}$. In Fig.~\ref{fig:error_vs_reg}
we visualize the quantile function $Q_{\textrm{err}}(p,R)$ defined
to satisfy 
\begin{equation}
P(|\hat{E}-E|<Q_{\textrm{err}}|R)=p,\label{eq:q_err}
\end{equation}
for various cumulative probabilities $p$ and non-hierarchicalities
$R$, combined over 8 random splits of the QM9 data using HIP-NN with
$N_{\textrm{feature}}=80$. In the background of the plot, we show
the distitribution of molecules using a histogram in $\log\,R$. The
bin width is $\Delta\log_{10}R=0.066$. The error of a random molecule,
drawn from a given bin of $R$, falls below the quantile $Q_{\textrm{err}}(p,R)$
with probability $p$; thus $1-p$ gives the empirical probability
that a molecular error will exceed $Q_{\textrm{err}}$.

We observe that, among the vast majority of the dataset ($R\gtrsim3\times10^{-2})$,
increasing $R$ corresponds to larger error quantiles. In other words,
if the energy contributions $\hat{E}_{i}^{n}$ are more hierarchical
in $n$ for a given molecule, then HIP-NN is more likely to be accurate.
This is true both for the typical ($p=0.5$) and outlier ($p=0.99$)
quantiles $Q_{\textrm{err}}$.

\subsection{MD Trajectories}

Here, we demonstrate that HIP-NN also performs well when trained on
energies obtained from molecular dynamics (MD) trajectories. We use
datasets generated by Schütt et. al~\cite{Schuett17} consisting
of MD trajectories for four molecules in vacuum: benzene, malonaldehyde,
salicylic acid, and toluene. The temperature is $T=500$~K. Energies
and forces were calculated using density-functional theory with the
PBE exchange-correlation potential~\cite{Perdew96}. Previous studies
on the Gradient Domain Machine Learning (GDML)~\cite{Chmiela17}
and SchNet~\cite{Schuett17a} models have also benchmarked on this
dataset.

We use training datasets of two sizes, $N_{\textrm{train}}=1\textrm{k}$
and 50k, randomly sampled from the full MD trajectory data. We use
an additional $N_{\textrm{validate}}=1\textrm{k}$ random conformations
for early stopping. The remaining conformations from each MD trajectory
comprise the test data. For the case with $N_{\textrm{train}}=1\textrm{k}$
conformers, we use a very simple HIP-NN model with $N_{\textrm{on-site}}=0$
and $N_{\textrm{feature}}=20$, which corresponds to about 10k model
parameters. When training on $N_{\textrm{train}}=50\textrm{k}$ conformers,
we instead use $N_{\textrm{on-site}}=3$ and $N_{\textrm{feature}}=40$,
which corresponds to about 59k model parameters. The resulting MAE
benchmarks are shown in Table~\ref{tab:traj_results}. When restricted
to training on \emph{only }energies, HIP-NN is comparable to or better
than the other models included in this benchmark. However, our current
implementation of HIP-NN does not train on forces. When SchNet and
GDML are trained with force data, they outperform HIP-NN. Extending
our model to force training is straightforward and will be reported
in future work.

Finally, we note that the energies in this dataset are only expressed
with a precision of 0.1~kcal/mol,\footnote{Retrieved from \url{http://quantum-machine.org/datasets/#md-datasets}}
which is comparable to many MAEs in Table~\ref{tab:traj_results}.
This suggests that lower MAEs may be possible with a more precise
dataset, especially with training set size $N_{\textrm{train}}=50\textrm{k}$.

\begin{table*}
\caption{\label{tab:traj_results}Accuracy of energy predictions for finite
temperature molecular conformers. We report the MAE in units of kcal/mol
for various training set sizes, model types, and molecule types. Including
forces in the training data significantly improves the predictions.
Best results for each training category are shown in bold.}

\begin{tabular}{lcc|cc||ccc|c}
\hline 
\multirow{2}{*}{} & \multicolumn{4}{c||}{$N_{\textrm{train}}=1\textrm{k}$} & \multicolumn{4}{c}{$N_{\textrm{train}}=50\textrm{k}$}\tabularnewline
 & \multicolumn{2}{c|}{Training on energy} & \multicolumn{2}{c||}{On energy \& forces} & \multicolumn{3}{c|}{Training on energy} & \multicolumn{1}{c}{On energy \& forces}\tabularnewline
 & HIP-NN & SchNet\cite{Schuett17a} & SchNet & GDML\cite{Chmiela17} & HIP-NN & SchNet & DTNN & SchNet\tabularnewline
\hline 
\hline 
Benzene  & $\mathbf{0.162}\pm0.002$ & $1.19$ & $0.08$ & \textbf{$\mathbf{0.07}$} & $0.064\pm0.002$ & $0.08$ & $\mathbf{0.04}$ & \textbf{$\mathbf{0.07}$}\tabularnewline
Malonaldehyde  & \textbf{$\mathbf{0.970}\pm0.019$} & $2.03$ & \textbf{$\mathbf{0.13}$} & $0.16$ & \textbf{$\mathbf{0.094}\pm0.001$} & $0.13$ & $0.19$ & \textbf{$\mathbf{0.08}$}\tabularnewline
Salicylic acid  & \textbf{$\mathbf{1.444}\pm0.024$} & $3.27$ & $0.20$ & \textbf{$\mathbf{0.12}$} & \textbf{$\mathbf{0.195}\pm0.002$} & $0.25$ & $0.41$ & \textbf{$\mathbf{0.10}$}\tabularnewline
Toluene  & \textbf{$\mathbf{0.880}\pm0.019$} & $2.95$ & \textbf{$\mathbf{0.12}$} & \textbf{$\mathbf{0.12}$} & \textbf{$\mathbf{0.144}\pm0.004$} & $0.16$ & $0.18$ & \textbf{$\mathbf{0.09}$}\tabularnewline
\hline 
\end{tabular}
\end{table*}

\section{Discussion}

\label{sec:discussion}

HIP-NN achieves state-of-the-art performance on both QM9 and MD trajectory
datasets, with MAEs well under 1 kcal/mol. We show that HIP-NN continues
to perform well even when the parameter count is drastically reduced.
We attribute the success of HIP-NN to a combination of design decisions.
One is the use of sensitivity functions~\cite{Duvenaud15,Schuett17,Schuett17a}
with an inverse-distance parameterization~\cite{Chmiela17,Han17,Montavon12,Hansen15}.
Thus we achieve finer sensitivity at shorter ranges and coarser sensitivity
at longer ranges. Another effective design decision is the use of
the ResNet transformation, Eq.~(\ref{eq:resnet}), a now-common technique
to improve deep neural networks~\cite{He16,Schuett17a}. A small
amount of $L_{2}$ regularization, Eq.~(\ref{eq:l2_regularization}),
is very helpful for stabilizing the root-mean-squared error, but has
little effect on the MAE. Annealing the learning rate when the validation
score plateaus improves optimization of the model parameters.

The physically motivated hierarchical energy decomposition, Eq.~(\ref{eq:energy_2}),
and corresponding regularization, Eq.~(\ref{eq:hierarchical_regularization}),
noticeably improve HIP-NN performance. Without this decomposition,
the MAE increases by 9\% and the fraction of errors under 1 kcal/mol
increases by 22\%. This improvement is intriguing, given that the
energy decomposition negligibly increases the total parameter count.
Also, the lower order energy contributions are formally redundant
given that the linear pass-throughs ($\tilde{M}_{ab}^{\ell}=\delta_{ab}$)
of the ResNet transformation, Eq.~(\ref{eq:resnet}), could allow
features to propagate unchanged through the network.

We interpret the hierarchical energy terms as follows. At zeroth order,
$\hat{E}_{i}^{n=0}$ corresponds to the dressed atom approximation~\cite{Hansen15}.
Next, $\hat{E}_{i}^{n=1}$ captures information about distances between
atom $i$ and its local neighbors, but goes beyond traditional pairwise-interactions
by combining local pairwise information. The final term, $\hat{E}_{i}^{n=2}$,
captures more detailed geometric information such as angles between
atom triples. For our best performing models with fixed $N_{\textrm{interaction}}=3$,
we find that the truncated model energy $\hat{E}_{\textrm{trunc}}^{k}=\sum_{i}\sum_{n=0}^{k}\hat{E}_{i}^{n}$
has an MAE that decays exponentially with $k$.

Despite achieving state-of-the-art MAEs, we still find that the HIP-NN
energy predictions on QM9 have an error exceeding 1~kcal/mol about
2.3\% of the time. For certain applications this error rate may not
be acceptable. Future work may focus on developing models that have
a lower failure rate. Another important research direction is to develop
methods for inferring when the model prediction is unreliable. We
provide a step in this direction by showing that large $R$ (which
indicates failure of the hierarchical energy decomposition) implies
that the HIP-NN energy prediction is less reliable.

As methodology improves, the machine learning community has room to
study increasingly challenging and varied datasets (e.g. Refs.~\onlinecite{Schuett17a,Smith17a})
in pursuit of improved accuracy and transferability. Other interesting
research directions include using active learning to construct diverse
datasets that cover unusual corners of chemical space~\cite{Li15,Huang17,Gubaev17,Podryabinkin17},
and using machine learning to glean chemical and physical insight~\cite{Yao17a}.

\section{Conclusion}

\label{sec:conclusion}

This paper introduces and pedagogically describes HIP-NN, a machine
learning technique for modeling molecular energies. By using an appropriate
molecular representation, HIP-NN naturally encodes permutation, rotation
and translation invariances. Inspired by the many-body expansion,
HIP-NN also encodes locality and hierarchical properties that one
would expect of molecular energies from physical principles. HIP-NN
improves significantly upon the state-of-the-art in predicting energies
on the QM9 dataset, a standard benchmark of organic molecules. HIP-NN
also shows promise on datasets of finite-temperature molecular trajectories.
The HIP-NN energy function is smooth, and thus can potentially drive
MD simulations. In addition to enhancing performance, the hierarchical
decomposition of energy yields an empirical measure of model uncertainty:
If the energy hierarchy produced by HIP-NN does not decay sufficiently
fast, the corresponding molecular energy prediction is less likely
to be accurate. 

\acknowledgments

The authors thank Benjamin Nebgen, Adrian Roitberg, and Sergei Tretiak
for valuable discussions and feedback. Our work was supported by the
Laboratory Directed Research and Development (LDRD) program, the Advanced
Simulation and Computing (ASC) program, and the Center for Nonlinear
Studies (CNLS) at Los Alamos National Laboratory (LANL). Computations
were performed using the CCS-7 Darwin cluster at LANL.

\bibliographystyle{apsrev4-1}
%\bibliography{refs}

\appendix
\end{document}